\documentclass{article} % For LaTeX2e
\usepackage{colm2024_conference}

\usepackage{microtype}
\usepackage{hyperref}
\usepackage{url}
\usepackage{booktabs}

\usepackage[most]{tcolorbox}
\usepackage{pifont}
\usepackage{colortbl}
\usepackage{CJKutf8}
\usepackage{multirow}
\usepackage{graphicx}
\usepackage{multicol}
\usepackage{xcolor}
\usepackage{wrapfig}
\usepackage{listings}
\usepackage{placeins}
\usepackage{tabularx}

\newcommand\bluecolor[1]{\cellcolor{gray!40!blue!30}{#1}}
\newcommand\greencolor[1]{\cellcolor{gray!40!green!30}{#1}}

\newcommand\purplecolor[1]{\cellcolor{gray!20!purple!30}{#1}}

\title{InternLM-Law: An Open Source Chinese Legal Large Language Model}

\author{Zhiwei Fei$^{1,2*}$, Songyang Zhang$^{1*}$, Xiaoyu Shen$^3$, Dawei Zhu$^4$, Xiao Wang$^4$, Maosong Cao$^1$,\\
\textbf{ Fengzhe Zhou$^1$, Yining Li$^1$, Wenwei Zhang$^1$, Dahua Lin$^1$, Kai Chen$^{1\dagger}$, Jidong Ge$^{2\dagger}$}
\\
\\
\textbf{$^1$Shanghai AI Laboratory}\\
\textbf{$^2$Nanjing University}\\
\textbf{$^3$Eastern Institute of Technology, Ningbo}\\
\textbf{$^4$Saarland University, Saarland Informatics Campus}\\
\texttt{internlm@pjlab.org.cn}\\
$^*$ Equal contribution\\$^\dagger$ Corresponding authors
}

\colmfinalcopy
\begin{document}

\maketitle
\begin{abstract}
While large language models (LLMs) have showcased impressive capabilities, they struggle with addressing legal queries due to the intricate complexities and specialized expertise required in the legal field. In this paper, we introduce InternLM-Law, a specialized LLM tailored for addressing diverse legal queries related to Chinese laws, spanning from responding to standard legal questions (e.g., legal exercises in textbooks) to analyzing complex real-world legal situations. 
We meticulously construct a dataset in the Chinese legal domain, encompassing over 1 million queries, and implement a data filtering and processing pipeline to ensure its diversity and quality. Our training approach involves a novel two-stage process: initially fine-tuning LLMs on both legal-specific and general-purpose content to equip the models with broad knowledge, followed by exclusive fine-tuning on high-quality legal data to enhance structured output generation. InternLM-Law achieves the highest average performance on LawBench, outperforming state-of-the-art models, including GPT-4, on 13 out of 20 subtasks. We make InternLM-Law and our dataset publicly available to facilitate future research in applying LLMs within the legal domain.\footnote{Our dataset, code and models will be released at https://github.com/InternLM/InternLM-Law.}
\end{abstract}

\section{Introduction}

Large Language Models (LLMs) are a significant research direction in the field of Natural Language Processing (NLP) and have attracted increasing attention from researchers~\cite{brown2020language,su2022rocbert,su2022welm,team2023internlm}.  Some studies are applying large language models to various fields such as medicine~\citep{singhal2023large}, coding~\citep{tang2021ast,muennighoff2023octopack}, mathematics~\citep{romera2024mathematical,ying2024internlmmath}, etc. These large language models can help solve domain-specific problems, and instruction-following models can respond using fluent natural language. In legal intelligence, some early studies have explored building legal domain-specific large language models to solve legal tasks. Previous works are mainly focused on providing legal advice~\citep{yue2023disc}, have limited application scenarios, 
and are based on early large language models  (e.g. Llama-1)~\citep{touvron2023llama}, which do not perform as well as some general large language models~\citep{fei2023lawbench}. Building a large legal model in the Chinese domain remains a worthwhile problem to explore.

\begin{figure}[!ht]
    \centering
    \includegraphics[width=1\columnwidth]{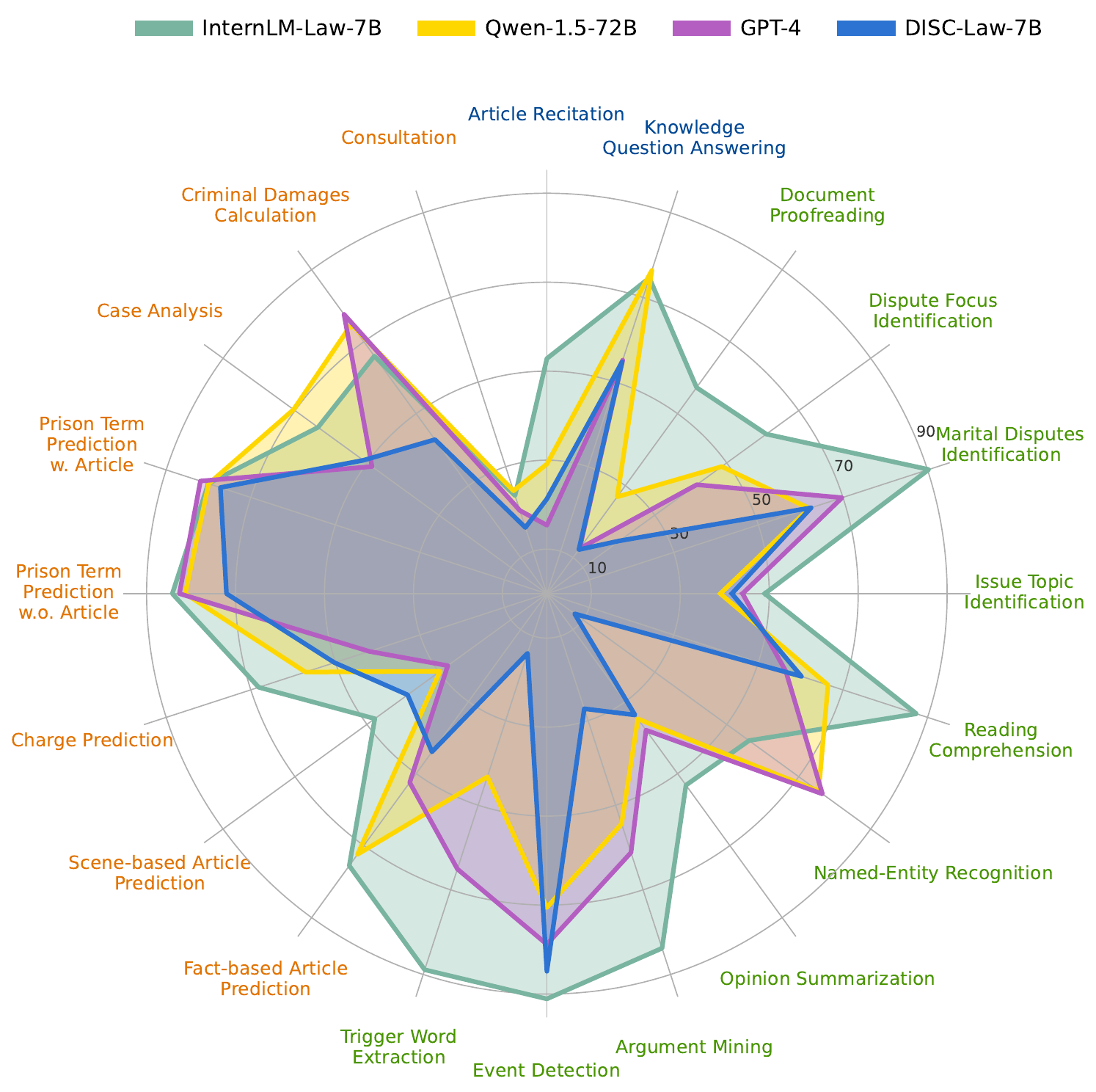}
    \caption{
    Model performance (zero-shot) evaluated across 20 legal subtasks from LawBench. Our proposed model, InternLM-Law-7B, achives the highest overall performance, outperforming GPT-4 and other larger-scale Chinese general-domain model such as Qwen-72b.}
    \label{figure1}
\end{figure}

In this study, we introduce \textbf{InternLM-Law}, a large language model tailored for the Chinese legal domain. We collect and clean a comprehensive supervised fine-tuning dataset sourced from various public legal datasets on the internet, encompassing both question-and-answer and non-Q\&A datasets. To convert diverse formats of data into supervised fine-tuning datasets, we have devised tailored data-cleaning pipelines, procedures, and augmentation strategies to uphold data quality and diversity. When solving legal issues, it is necessary to rely on some general abilities such as text understanding and analytical skills. Merely relying on legal datasets to imbue the model with legal capabilities is inadequate. To enable the model to transfer general skills to solving legal tasks, we incorporate SFT data sampled from the training of the InternLM2-Chat model and trained it in conjunction with legal datasets. Additionally, we design a two-stage SFT training strategy to enhance the model's ability to effectively learn crucial datasets, such as laws and regulations, and to adjust the model's response style for an optimal subjective experience.

Our contributions can be summarized as follows:
\begin{itemize}
\item We develop InternLM-law, a LLM specifically designed for the Chinese legal domain. This model is capable of performing a broad spectrum of legal tasks, establishing a new state-of-the-art (SOTA) on the LawBench benchamrk. It outperforms all existing publicly available legal models and larger general-purpose models like GPT-4.

\item To build this strong LLM, we devote extensive efforts to both data collection and training stages. In particular, we build a large scale dataset tailored for the Chinese legal domain, containing over 1M training samples. We implement effective data filtering and processing techniques to ensure the dataset's quality. We train InternLM-law using a novel, two-stage SFT pipeline; we frist train it on both legal-related and general domain tasks, followed by high-quality legal-specific data.

\item Beyond standard benchmark, we enrich our evaluation with subjective and long-text evaluation. These supplementary evaluation metrics are intended to augment the existing Chinese judicial assessment framework.
\end{itemize}

\section{Related Work}

Legal Artificial Intelligence (LegalAI)~\citep{zhong2020does} has been a long-standing research problem in the field of Natural Language Processing (NLP). Previous works mainly focused on building special models for one task (e.g. legal judgment prediction~\citep{ge2021learning,huang2021dependency,cui2023survey}). These independent tools make a legal system complex. Some researchers are exploring developing domain-specific large language models tailored for the legal domain to various scenarios. There have been some initial attempts at building legal LLMs. SaulLM-7B~\citep{colombo2024saullm} is a legal text model based on the Mistral 7B architecture, trained on a 30 billion-token English legal corpus, designed explicitly for legal text comprehension and generation. In the Chinese legal domain, Lawyer llama~\citep{huang2023lawyer} explores how to build a large language model. They further pretraining the Chinese-llama-13B on legal datasets and enhance its consulting capabilities through further fine-tuning on legal consultation-related datasets. ChatLaw~\citep{cui2023chatlaw} explores the impact of model size on model performance. They trained two large legal language models on Ziya-LLaMA-13B-v1~\citep{fengshenbang} and Anima-33B\footnote{https://huggingface.co/lyogavin/Anima33B} and found that a larger-scale base could enhance the model's analytical capabilities. These models focus solely on legal consultations, while the DISC-Law~\citep{yue2023disc} model further expands its application scenarios, enabling the model to provide a wider range of services such as completing legal tasks, legal consultations, and legal exam assistance. However, they only tested this model on legal exams and not for other capabilities.

\section{Training Process}

\begin{figure*}[htbp]
    \centering
    \includegraphics[scale=0.25]{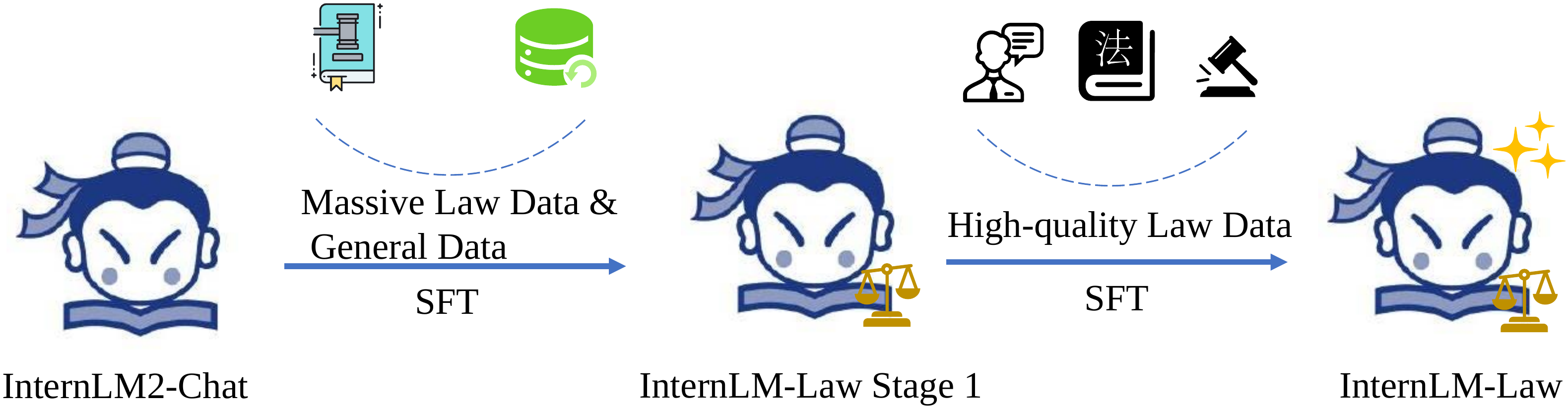}
    \caption{Illustration of our two-stage training pipeline. Initially, InternLM2-chat is trained on a diverse range of tasks including both general-purpose and legal tasks. Subsequently, it is further trained specifically on high-quality legal tasks.}
    \label{figure4}
\end{figure*}

We employ InternLM2-Chat as our foundation model and perform a two-stage supervised fine-tuning (SFT) to specialize it for legal domain. The training pipeline is presented in Figure \ref{figure4}. In the first stage, we train the model on a mixture of legal and general-purpose tasks. This steps aims to enrich the model's understanding of the Chinese legal domain compromising its versatility across general domains. Empirically, we find that incorporating general-purpose tasks helps improve the model's ability in the legal domain. In the second stage, we refine the model by conducting additional training on high-quality legal tasks (detailed in Section~\ref{Section54}), enhancing its response style, answering structure, and factual accuracy in addressing legal inquiries. Section~\ref{sec:ablation-study} provides a comprehensive ablation study that disentangles the effectiveness of each training stage.

We train our model with 64 A100-80GB for 8 hours. To enable the model to process long legal texts, we set the training sequence length to 32k, ensuring that our model can accommodate most legal text inputs. During training, we set the learning rate to 1e-5, and each stage was trained for one epoch.

\section{Data Sources} \label{data_sources}

Our data sources are composed of two parts: the legal domain dataset and the general domain dataset. In this section, we will provide details of the origins and composition of these two parts.

\subsection{Legal Data Sources}
Our objective is to construct a comprehensive dataset encompassing a broad spectrum of legal knowledge. This dataset is organized into three main categories: legal NLP, legal consultation, and legal regulations. The first two categories encompass a variety of tasks related to legal education and practice, including preparation materials for legal examinations and realistic scenarios for the application of legal knowledge. The final category is comprised of up-to-date legal regulations, essential for teaching LLMs accurate legal knowledge. This enhances factual accuracy and minimizes the generation of erroneous or fabricated information by LLMs.
\paragraph{Legal NLP Data.}

Legal NLP data comprises tasks within the legal domain that are well-defined and exhibit consistent input and output formats, akin to standard NLP tasks. This data is sourced primarily from prior research endeavors in the field, including datasets from public legal competitions like CAIL~\footnote{http://cail.cipsc.org.cn/}, training data provided by LawBench~\citep{fei2023lawbench}, legal information extraction~\citep{yao-etal-2022-leven}, legal judgment prediction frameworks~\citep{xiao2018cail2018}, and others. In aggregate, we have curated 22 distinct legal-related tasks, yielding a comprehensive dataset consisting of 440K samples.

\paragraph{Legal Consultation Data.}

We collected a massive amount of legal consultation data, consisting of 6 million records obtained from various online platforms. These records represent a broad spectrum of real-world legal issues, spanning civil disputes, policy interpretation, and criminal cases. The data primarily consists of queries posed by individuals, which are then responded to by experienced legal practitioners, thus creating a rich collection of question-and-answer pairs. To ensure confidentiality, robust anonymization procedures were carried out to safeguard sensitive information.

\paragraph{Chinese laws \& Regulations Data.}
The foundational premise for a Large Language Model (LLM) to accurately address legal issues is its ability to incorporate relevant laws and regulations. To this end, we have sourced legal regulations data from the Chinese National Legal Database\footnote{https://flk.npc.gov.cn/}, a comprehensive data source encompassing Chinese civil, criminal, and constitutional law, along with an extensive range of regulations. This endeavor aims to infuse the LLM with with precise and authoritative legal information. In total, we collected 100K entries from this database.

\subsection{General Data Sources}
We follow the recent works to build the general SFT data. In the supervised fine-tuning (SFT) stage, we use a dataset of 1 million instruction data instances, which have been screened to ensure their helpfulness and harmlessness. Our dataset presents a comprehensive and eclectic array of subject matters, encompassing everyday conversation, natural language processing (NLP) tasks, mathematical problems, code generation, function callings, and more. In order to maintain consistency and comparability, we adhere to the identical topic distribution found in InternLM2. Recognizing the inherent complexity and diversity of these tasks, we judiciously employ the ChatML framework~\citep{ChatML} to uniformly structure and represent each data sample, thereby fostering adaptability and ease of processing across this multifaceted collection.

\section{Legal Data Processing}

We devise a detailed data processing plan for the dataset mentioned in Section~\ref{data_sources}. Further, we observe that legal consulting datasets crawled from websites often provide overly brief and insufficiently detailed responses. To address this issue, we develop a semi-automated paraphrasing method to generate more comprehensive and high-quality datasets. The distribution of data in these datasets is highly imbalanced, we sampled some key datasets from the processed data, including laws \& regulations, well-written Q\&A, and various legal tasks, to create a high-quality legal dataset. Detailed statistics information can be found in Appendix ~\ref{AppendixA}.

\begin{figure*}[!t]
    \centering
    \includegraphics[scale=0.27]{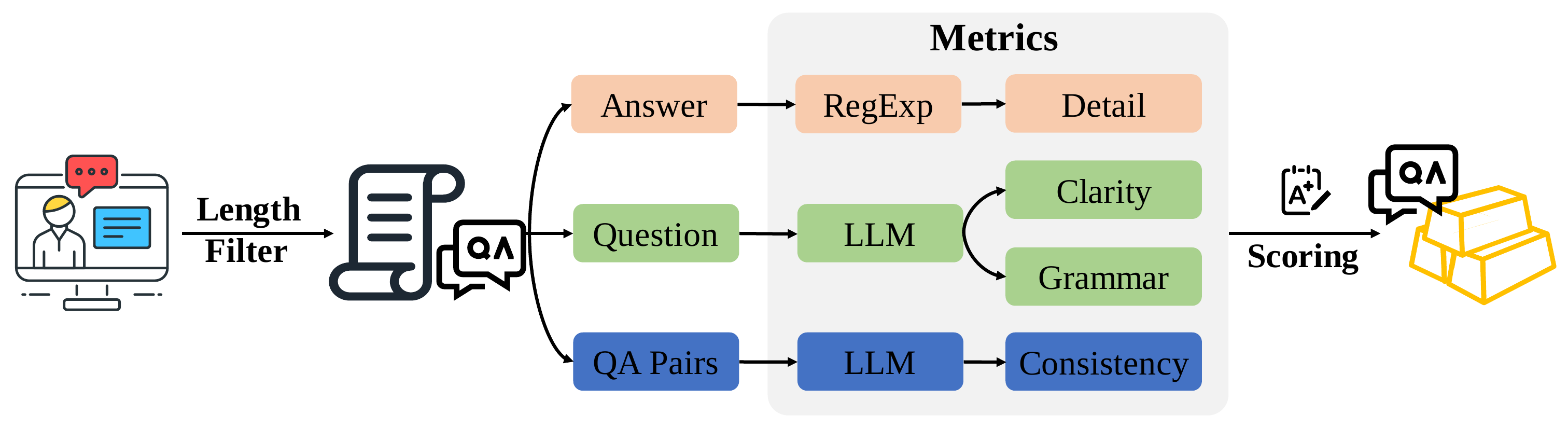}
    \caption{
    Data processing pipeline for legal consultation data. Initial filtering of answers is conducted using rule-based methods, such as regular expressions. Subsequently, the quality assessment of questions and the coherence between questions and answers is performed utilizing LLMs.}
    \label{legal-consultation-illustration}
\end{figure*}

\subsection{Legal NLP Data Processing}
When dealing with structured NLP tasks related to law, we refer to LawBench to categorize these legal tasks into five types: single-label classification, multi-label classification, extraction, generation, and regression. We first manually write instructions for each legal task. Then use GPT-4 to generate diversified and semantically similar instructions for each task. We randomly choose one of these instructions from the set and use it to construct a legal task dataset, enhancing the diversity of the legal task dataset. The detailed process is described in Figure \ref{figure1}.

\subsection{Legal Consultation Data Processing}
Our legal consultation dataset, compiled from a wide range of  online sources, encompasses a wide range of legal scenarios. However, it also contains extraneous information that can detrimentally affect the quality of our data. To ensure the reliability of our training data, we use a series of filtering methods to refine the dataset. An overview of this filtering process is shown in Figure~\ref{legal-consultation-illustration}.

\paragraph{Rule-based Filtering.}

Recognizing the complexity of legal consultations, which often involve complex legal matters requring thorough analysis and discussion, comprehensive answers should exceed a certain length. Hence, we discard answers less than 20 characters for lacking necessary detail. Additionally, high-quality answers typically reference legal provisions, indicated by \begin{CJK}{UTF8}{gbsn}``《''\end{CJK} and \begin{CJK}{UTF8}{gbsn}``法''\end{CJK} in formal Chinese writing. Responses lacking these markers are filtered out. This strategy yields a refined dataset of 1 million legal consultations.

\begin{wrapfigure}[23]{r}{0cm}
    \centering
    \includegraphics[scale=0.25]{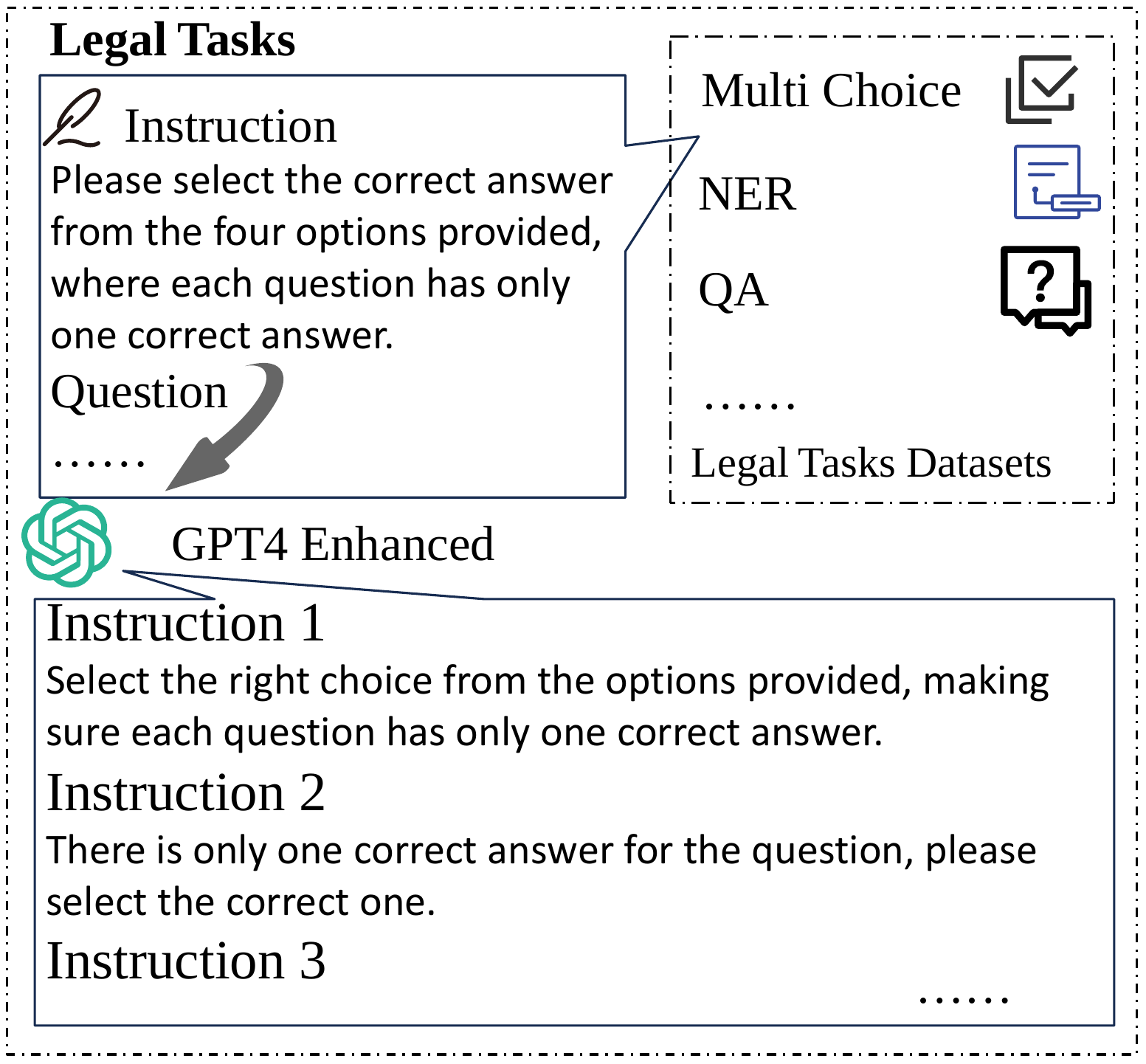}
    \caption{
    We manually create a set of seed instructions for each task, which are subsequently expanded using GPT-4 through paraphrasing, resulting in a rich pool of diverse instructions. During training, each training instance is paired with an instruction randomly chosen from this pool.}
    \label{figure1}
\end{wrapfigure}

\paragraph{Semantic Filtering.}
After filtering out instances with potentially inferior answers, our focus turns to instances characterized by poorly formulated questions. Notably, we notice that questions may lack clarity and contain grammar errors, which should be pruned. To address this issue, we leverage a LLM to identify such instances. This model assesses instances based on the clarity and correctness of their questions, discarding those that fall below a predetermined threshold. Subsequently, we evaluate the coherence between questions and answers, employing the same LLM for this task. It checkes whether questions are comprehensively understood and appropriately addressed, filtering out instances that receive low scores. Throughout this process,  the LLM we select is Qwen-1.5-72B\footnote{https://huggingface.co/Qwen/Qwen1.5-72B}, because it is one of the strongest LLMs in the Chinese domain\footnote{https://qwenlm.github.io/blog/qwen1.5/}.

\subsection{Chinese laws \& Regulations Processing}
Since the legal regulation dataset is pure text format. We transform dataset into question-answering pairs for supervised fine-tuning. By extracting the titles of laws or regulations, we design different prompts to the titles, and convert them into questions, which requires models to recite the content of different laws or regulations. And the answer is the text of corresponding laws or regulations. 
The approach can encode legal knowledge into the model, enabling model to memorize the exact content of laws.
\subsection{High-quality Legal Data Processing} \label{Section54}
These datasets are utilized to refine the model's conversational style and to render its legal knowledge more precise. To achieve this, we utilized GPT-4 to generate high-quality Q\&A datasets semi-automatically. The generated content was then manually adjusted for precision. Furthermore, we resampled crucial data from previously supervised fine-tuning datasets.

\paragraph{Data Synthesis.}
There are significant inconsistencies in the style of datasets written by humans, with some of the content being very short. We synthesized some data using GPT-4 combined with human feedback to unify the style of responses and make them more detailed. The synthesis of data involves three steps. Firstly, we selected 6K question-answer pairs crawled from the internet, which consists of questions posed by individuals and brief responses manually written, including references to laws and regulations. We engaged three legal background annotators to further expand on these questions from different perspectives and incorporate relevant legal references. Subsequently, we prompted the information and feed into GPT-4 which can generate more detailed responses. By manually checking the logic of the responses and the accuracy of the referenced content, correcting any inaccuracies. We got a high-quality dataset of detailed and stylistically consistent answers.

\paragraph{Data Resampling.}
To minimize model hallucinations in interpreting laws and regulations while preserving its competency in other legal tasks, we primarily selected important laws \& regulations and tasks for sampling.

\textit{Resampling important law:} Based on the distribution of legal content extracted from our web-scraped legal question-answer pairs, we resampled the most frequently occurring crucial legal content, such as marriage law, labor law, criminal law, constitutional law, etc., from the Chinese Civil Code. These provisions enhance the model's accuracy in generating legal provisions when responding to legal content.

\textit{Resampling Legal NLP tasks:} To preserve the legal task capabilities of the model learned from stage-one SFT, we resampled past legal datasets to ensure that the model retains its proficiency in legal task handling.

\section{Experiments}

In this section, we compare our model with the SOTA models on various datasets, covering multiple evaluation methods including legal benchmark evaluation, subjective evaluation, and long-text evaluation. We illustrate some examples of applications in Appendix \ref{AppendixC}.
\subsection{Model Baselines}
We compare our model with strong LLM baselines from different groups:

\textbf{General-purpose LLMs:} We select the latest advanced LLMs that are pre-trained on massive data in Chinese domain, including Qwen-1.5-7B-Chat ~\citep{bai2023qwen}, InternLM2-7B-Chat ~\citep{team2023internlm}, and Qwen-1.5-72B-Chat.

\textbf{Legal-specific LLMs:} This group comprises LLMs specifically tailored for Chinese legal applications, including DISC-Law-7B~\citep{yue2023disc}, Lawyer-LLaMA-13B~\citep{huang2023lawyer}, ChatLaw-13B~\citep{cui2023chatlaw}, and fuzimingcha-6B~\citep{sdu_fuzi_mingcha}.

In addition, we include GPT-4, a strong commercial LLM that has leading position on various leaderboards.

\subsection{Legal Benchmark Evaluation}
\paragraph{Evaluation Setup.}
We evaluate LLMs in legal contexts using LawBench~\citep{fei2023lawbench}, a well-established benchmark tailored for the Chinese legal domain. It assesses models on memorization, understanding, and application across 20 tasks. In addition to direct evaluation (zero-shot), we report the model's one-shot performance, to take into account cases where LLMs are not familiar with the input-output format of a given task. For decoding, we set a maximum generation length of 1024 and use greedy decoding.

\begin{table*}[!thb]
\centering
\small
\label{tab: bilingual score}
\begin{tabular}{lcccc}
\hline
\textbf{Models} & \textbf{Memorization} & \textbf{Understanding} & \textbf{Application} &\textbf{Average} \\
\hline

\multicolumn{5}{c}{\textit{\purplecolor{Commercial LLMs}}} \\
GPT-4 & 35.29/36.01 & 54.41/56.48 & 54.05/55.01 & 52.35/53.85 \\  
\hline
\multicolumn{5}{c}{\textit{\bluecolor{General-purpose LLMs}}} \\
Qwen-1.5-7B & 19.16/19.00 & 39.19/43.62 & 49.75/50.40 & 41.41/43.87 \\
InternLM2-7B & 31.62/32.02 & 45.00/48.33 & 50.33/52.91 & 45.80/48.53 \\
Qwen-1.5-72B & 52.77/50.06 & 52.16/54.92 & 61.24/62.28 & 55.85/57.38 \\
\hline
\multicolumn{5}{c}{\textit{\greencolor{Legal-specific LLMs}}} \\
Lawyer-LLaMA-13B& 17.77/11.82 & 18.94/18.89 & 35.19/30.99 & 25.32/23.02 \\
ChatLaw-13B& 21.63/22.69 & 28.21/30.22 & 41.23/38.13 & 32.76/32.63 \\
Fuzimingcha-6B& 16.51/16.51 & 30.10/23.83 & 40.86/38.04 & 33.05/28.78 \\
DISC-Law-7B & 38.05/37.02 & 36.43/38.07 & 48.94/53.14 & 41.60/43.99 \\
InternLM-Law-7B \textit{(Ours)}& \textbf{63.72}/\textbf{64.95} &\textbf{ 71.81}/\textbf{71.58} & 
\textbf{63.57}/\textbf{63.46} & \textbf{67.71}/\textbf{67.67} \\
\hline
\end{tabular}
\caption{Model performance on LawBench. The numbers represent the zero-shot/one-shot performance.}
\label{tab:model-performance-on-lawbench}
\end{table*}

\paragraph{Results.}

The performance metrics are shown in Table ~\ref{tab:model-performance-on-lawbench}. GPT-4 demonstrates non-trivial performance on LawBench, substantially outperforming smaller, general-purpose LLMs like the 7B variants of Qwen and InternLM2, which are trained on extensive data in both English and Chinese. However, the larger Qwen-72B model, which also focuses on English and Chinese, outperforms GPT-4. Interestingly, legal-specific LLMs do not always exceed the performance of general-purpose models, often due to their struggles with instruction compliance. This highlights the importance of integrating instruction-following skills with legal knowledge in LLM training. InternLM-Law, in contrast, shows state-of-the-art performance, outperforming all competitors, including GPT-4, across all assessed dimensions.\footnote{Each dimension's performance is averaged across all associated tasks. For task-specific details, refer to Appendix \ref{appx:sec:lawbench-detailed-results}.} Notably, InternLM-Law also significantly advances over its base model, InternLM2-7B, across all tasks, showcasing effective knowledge transfer through our two-stage training process.

\subsection{Subjective Evaluation}
\paragraph{Evaluation Setup.}

In addition to benchmark evaluation, we assess the ability of LLMs to respond to subjective legal questions, mirroring a real-world legal consultation scenario. For this analysis, we carefully bulit a dataset of 100 questions, categorized into three types: legal consultation, case analysis, and legal reasoning, distributed in a ratio of 40/30/30. For each question, we compare the responses of LLMs with those provided by GPT-4. Given the open-ended nature of subjective questions, we use GPT-4 as a judge to gauge other LLMs' performance, reporting their win rates against GPT-4’s responses.

To avoid position bias~\citep{zheng2024judging} and compare the model's performance based on semantics rather than location, we swapped the model's output and used GPT-4 to judge which model performed better in two different tests. If GPT-4 selected different models in the tests, we would discard the result. Only when the model consistently provided the same results for separate tests that differed in position, did we calculate the model's win rate based on the consistency in model judgments.

\paragraph{Results.}
As can be seen in Table \ref{table2}, our model achieves an average result of 46.67\% win-rate across the three tasks, on par with GPT-4. Outperforms GPT-4 on the legal consultation task, achieving a win-rate of 87.5\%. Earlier legal models like ChatLaw have the worst performance due to their lack of training in subjective tasks.

\begin{table}[ht]
\centering
\resizebox{\columnwidth }{!}{
\begin{tabular}{lccccc} 
\toprule
\textbf{Models} & \textbf{Overall}  & \textbf{Legal Consultation} & \textbf{Case Analysis} & \textbf{Legal Reasoning}& \textbf{Position Bias}\\
\midrule
Lawyer-LLaMA & \phantom{0}0.0 & \phantom{0}0.0  & \phantom{0}0.0 & \phantom{0}0.0 & 0\\
ChatLaw & \phantom{0}0.0 & \phantom{0}0.0  & \phantom{0}0.0 & \phantom{0}0.0  & 0 \\
Fuzimingcha & \phantom{0}5.0 & 10.0  & \phantom{0}0.0& \phantom{0}5.0 & 1\\
DISC-Law & 11.7 & 25.0   & 10.0& \phantom{0}0.0 & 3\\
InternLM-Law Stage 1 & \phantom{0}7.5 & 10.0  & 12.5 & \phantom{0}0.0 & 2\\
InternLM-Law & 46.7 & 87.5  & 32.5& 20.0 & 5 \\
\bottomrule
\end{tabular}}
\caption{The model's performance on the legal consultation task is superior to GPT-4, based on objective evaluation.}
\label{table2}
\end{table}

\subsection{Long Context Evaluation}
\paragraph{Evaluation Setup.} 
In legal tasks, handling with long legal judgment is common~\citep{xiao2021lawformer}. Previous work shows limitations of BERT~\citep{devlin2018bert} and other models in dealing with long legal judgment. We created a dataset focusing on analyzing Chinese law judgments. It includes 20 legal judgments over 20k characters each, with three associated questions per judgment. The model must accurately locate relevant content for each question through reading comprehension, without multi-hop reasoning. We evaluate the model's ability to recall information from the legal judgments. LMDeploy~\citep{2023lmdeploy} was used as the inference backend, with the input length set to 25000.

\paragraph{Results.}

Early models, such as Lawyer-LLaMA, ChatLaw, and fuzimingcha, are unable to take extremely long text as input, leading to their failure on our test. Other large legal models, such as DISC-Law, which are based on newer models, can accept long text but struggle to handle complex information. It achieves an F1 score of 36.72\% on our task. In contrast, our model can effectively process long text information and retrieve the necessary information from legal documents and achieves 84.73\% F1 score.

\subsection{Ablation Studies}
\label{sec:ablation-study}
In this section, we delve into two aspects in detail. The first aspect is the impact of incorporating general datasets during training on legal tasks and general tasks. Subsequently, we elaborate on the role of two-stage SFT, with an intuitive approach being the joint training of the two stages of SFT. We compare the results obtained from training using only the first stage with the results obtained from training by mixing the two stages together.

\paragraph{Effectiveness of General Datasets.}
We test the impact of incrementally increasing the proportion of general training data on the model's performance on legal and general tasks. We first test the performance of the InternLM2-Chat model on legal and general tasks. We selected five general tasks for evaluating the model's capabilities: CMMLU~\citep{li2023cmmlu}, C3~\citep{sun2020investigating}, CSL~\citep{li2022csl}, CMNLI~\citep{xu2020clue}, and CHID~\citep{zheng2019chid}. CMMLU was used to measure the model's domain ability, CHID was used to measure the model's linguistic ability, C3 and CSL were used to measure the model's comprehension ability, and CMNLI was used to measure the model's reasoning ability.

\begin{table}[ht]
\centering
\small
\begin{tabular}{lccccc} 
\toprule
\textbf{Models} & \textbf{CMMLU(F1)} & \textbf{C3(F1)} & \textbf{CSL(F1)}& \textbf{CMNLI(Acc)} & \textbf{CHID(F1)}\\
\midrule
InternLM-Chat & \textbf{63.03} & \textbf{91.43} & \textbf{50.00} & \textbf{71.95} & \textbf{77.72}\\
InternLM-Law w.o. & 44.29 & 82.52 & 49.38 & 59.61 & 41.58\\
InternLM-Law (10\%) & 59.23 & 89.75 & 50.00 & 63.10 & 73.76 \\
InternLM-Law (20\%) & 60.79 & 90.14 & 49.38 & 68.50 & 74.26  \\
\bottomrule
\end{tabular}

\caption{Impact of varying amounts of general data in the training set on general task performance. ``w.o.'' indicates without general datasets, while ``10\%'' refers to a 1:10 ratio of general data to legal data. }
\label{table3}
\end{table}

\begin{table*}[!thb]
\centering
\small
\label{tab: bilingual score}
\begin{tabular}{lcccc}
\hline
\textbf{Models} & \textbf{Memorization} & \textbf{Understanding} & \textbf{Application} &\textbf{Average} \\
\hline
InternLM-Chat & 31.62/32.02 & 45.00/48.33 & 50.33/52.91 & 45.80/48.53 \\
InternLM-Law w.o. & 54.22/54.63 & 72.12/72.54 & 63.78/63.89 & 67.32/67.57 \\
InternLM-Law (10\%) & 54.62/55.72 & 72.63/72.87 & 63.85/63.91 & 67.41/67.75 \\
InternLM-Law (20\%)  & \textbf{55.73}/\textbf{56.82} &\textbf{73.56}/\textbf{73.78} & 
\textbf{64.45}/\textbf{64.05} & \textbf{68.13}/\textbf{68.19} \\

\hline
\end{tabular}
\caption{Impact of datasets with different proportions of general data in the training set on the performance of legal tasks. The results refer to zero-shot/one-shot scores on LawBench.}
\label{table4}
\end{table*}

As can be seen from Tables \ref{table3} and \ref{table4}, the addition of a general dataset not only preserves the model's general ability but also further enhances its legal ability. We speculate that this may be because the general dataset allows the model to generalize some problem-solving capabilities to legal tasks. At the same time, we found that SFT training reduces the general competence of the base model, and adding more general data can gradually improve the model's general competence.

\paragraph{Effectivenss of two-stage SFT.}
The two-stage SFT approach is designed to improve the reply style and reduce the hallucination of the model on laws and regulations. We compare the results of training with only stage-one SFT on objective and subjective evaluations. We further analyze the results of combining the two stages.

\begin{table*}[!thb]
\centering
\small
\label{tab: bilingual score}
\resizebox{\columnwidth}{!}{
\begin{tabular}{lcccc}
\hline
\textbf{Models} & \textbf{Memorization} & \textbf{Understanding} & \textbf{Application} &\textbf{Average} \\
\hline
InternLM-Law (Stage 1) & 55.73/56.82 & \textbf{73.56}/\textbf{73.78} & \textbf{64.45}/\textbf{64.05} & \textbf{68.13}/\textbf{68.19} \\
InternLM-Law (Merged) & 55.62/56.33 & 73.12/73.45 & 65.02/64.36 & 67.89/68.02 \\
InternLM-Law (Two-Stage) & \textbf{63.72}/\textbf{64.95} & 71.81/71.58 & 63.57/63.46 & 67.71/67.67 \\
\hline
\end{tabular}
}
\caption{Effects of different training strategies on LawBench. The results refer to zero-shot/one-shot scores on LawBench.}
\label{table5}
\end{table*}

From the experimental results shown in Table \ref{table5}, we can see that firstly training with only one-stage SFT achieves very high scores on the objective legal task, but performs poorly on the subjective evaluation, where we find that its win-rate against GPT-4 is lower than that of the two-stage SFT as results shown in Table \ref{table2}. Secondly, two-stage SFT degrades on some legal tasks. However, the model demonstrates a significant improvement on tasks related to legal knowledge. Combining the two stages does not substantially improve objective performance nor does it affect subjective performance. This indicates that mixing a small number of high-quality data into the entire data set fails to highlight their contributions.

\section{Conclusion and Limitations}
In this paper, we introduce a large language model in the Chinese legal domain. We provide a detailed description of our data composition and processing pipeline. These methods can semi-automatically transform collected data into training datasets. We employ a two-stage SFT training approach to enhance the model's accuracy in legal facts and response style. We evaluate our model on the legal dataset, LawBench. Currently, testing in the Chinese legal domain is limited, so we supplement it with subjective legal evaluations and long-text evaluations. Extensive experimental results demonstrate that our model is currently state-of-the-art in the field, outperforming all publicly available judicial models, general models, and commercial GPT-4 models.

Although we have made every effort to reduce model hallucinations, our model, like other large language models, still produces hallucinations and inevitably generates some inaccurate responses. In addition, due to the relatively small model size, there is room for improvement on more complex legal reasoning tasks.

\bibliography{colm2024_conference}
\bibliographystyle{colm2024_conference}

\appendix
\section{Static Information of Legal Datasets} \label{AppendixA}
\begin{table*}[!h]
\centering
\small
\resizebox{\columnwidth}{!}{
\begin{tabular}{cccc} 
\toprule
Dataset & Task & Size & Source \cr
\midrule
\multirow{22}*{Legal Task Datasets} & Article Recitation & 20K & \href{https://flk.npc.gov.cn/}{FLK}\cr
 & Knowledge Question Answering & 20K & \href{https://jecqa.thunlp.org/}{JEC\_QA} \cr
 & Document Proofreading & 20K & \href{http://cail.cipsc.org.cn/task_summit.html?raceID=2\&cail_tag=2022}{CAIL2022} \cr
 & Dispute Focus Identification & 20K & \href{https://laic.cjbdi.com/}{LAIC2021} \cr
 & Marital Disputes Identification & 20K & \href{https://aistudio.baidu.com/datasetdetail/181754}{AIStudio} \cr
 & Issue Topic Identification & 20K & \href{https://github.com/liuhuanyong/CrimeKgAssitant}{CrimeKgAssitant} \cr
 & Reading Comprehension & 20K & \href{http://cail.cipsc.org.cn/task_summit.html?raceID=1\&cail_tag=2019}{CAIL2019} \cr
 & Named-Entity Recognition & 20K & \href{https://github.com/china-ai-law-challenge/CAIL2021/tree/main/xxcq}{CAIL2021} \cr
 & Opinion Summarization & 20K & \href{http://cail.cipsc.org.cn/task_summit.html?raceID=4\&cail_tag=2022}{CAIL2022} \cr
  & Argument Mining & 20K & \href{http://cail.cipsc.org.cn/task_summit.html?raceID=5\&cail_tag=2022}{CAIL2022} \cr
 & Event Detection & 20K & \href{https://github.com/thunlp/LEVEN}{LEVEN} \cr
 & Trigger Word Extraction & 20K & \href{https://github.com/thunlp/LEVEN}{LEVEN} \cr
 & Fact-based Article Prediction & 20K & \href{https://github.com/china-ai-law-challenge/cail2018}{CAIL2018} \cr
 & Scene-based Article Prediction & 20K & \href{https://github.com/LiuHC0428/LAW-GPT}{LawGPT} \cr
 & Charge Prediction & 20K & \href{https://github.com/china-ai-law-challenge/cail2018}{CAIL2018} \cr
 & Prison Term Prediction w.o. Article & 20K & \href{https://github.com/china-ai-law-challenge/cail2018}{CAIL2018}\cr
 & Prison Term Prediction w. Article & 20K & \href{https://github.com/china-ai-law-challenge/cail2018}{CAIL2018} \cr
 & Case Analysis & 20K & \href{https://jecqa.thunlp.org/}{JEC\_QA} \cr
 & Criminal Damages Calculation & 20K & \href{https://laic.cjbdi.com/}{LAIC2021} \cr
 & Consultation & 20K & \href{https://www.66law.cn/}{hualv.com} \cr
 & Judgement Generation & 20K & \href{https://aclanthology.org/2020.emnlp-main.56.pdf}{AC\_NLG} \cr
 & Named-Entity Recognition & 20K & \href{https://github.com/thulawtech/leec}{LEEC} \cr
\midrule
\multirow{1}*{Legal Consultation Datasets} & Legal Consultation & 500K & Web Scraping \cr
\midrule
\multirow{1}*{Chinese laws \& Regulations Datasets} & Legal Knowledge Embedding & 100K & Web Scraping \cr
\midrule
Total & & 1.04M & \cr
\bottomrule
\end{tabular}
}
\caption{\small Statistics of the first-stage legal SFT dataset.}
\label{table7}
\end{table*}

\begin{table*}[!h]
\centering
\small
\begin{tabular}{ccc} 
\toprule
Dataset & Task & Size  \cr
\midrule
\multirow{1}*{High-quality Consultation Datasets} & Legal Consultation & 6K \cr
\midrule
\multirow{1}*{Sampled Laws \& Regulations Datasets} & Legal Knowledge Embedding & 10K  \cr
\midrule
\multirow{1}*{Sampled Legal NLP Task Datasets} & Legal Tasks & 4K  \cr
\midrule
Total & & 20k \cr
\bottomrule
\label{}
\end{tabular}
\caption{\small Statistics of the high-quality SFT dataset.}
\label{table8}
\end{table*}
\FloatBarrier

\section{Detailed Results on LawBench} \label{appx:sec:lawbench-detailed-results}

There are three categories of tasks in LawBench, with each category corresponding to one cognitive level, and indicating the model's capacity to deal with legal knowledge tasks. Table \ref{tab:table8} 

\begingroup
\begin{table*}[ht]
\centering
\small
\resizebox{\columnwidth}{!}{
\begin{tabularx}{1.1\textwidth}{p{0.7cm}p{11.5cm}X}
\toprule
\textbf{ID} & \textbf{Definition} & \textbf{Type} \\
\midrule
1-1 & {Article Recitation: Given a law article number, recite the article content.} & {Generation} \\\midrule
1-2 & {Knowledge Question Answering: Given a question asking about basic legal knowledge, select the correct answer from 4 candidates.} & {SLC}  \\\midrule
2-1 & {Document Proofreading: Given a sentence extracted from legal documents, correct its spelling, grammar and ordering mistakes, return the corrected sentence} & Generation \\\midrule
2-2 & Dispute Focus Identification: Given the original claims and responses of the plaintiff and defendant, detect the points of dispute. & MLC \\\midrule
2-3 & Marital Disputes Identification: Given a sentence describing marital disputes, classify it into one of the 20 pre-defined dispute types. & MLC \\\midrule
2-4  & Issue Topic Identification: Given a user inquiry, assign it into one of pre-defined topics. &  SLC \\\midrule
2-5 & Reading Comprehension: Given a judgement document and a corresponding question, extract relevant content from it to answer the question. & Extraction  \\\midrule
2-6 & Named-Entity Recognition: Given a sentence from a judgement document, extract entity information corresponding to a set of pre-defined entity types such as suspect, victim or evidence. &  Extraction \\\midrule
2-7 & Opinion Summarization: Given a legal-related public news report, generate a concise summary. &  Generation \\\midrule
2-8 & Argument Mining: Given a plaintiff's perspective and five candidate defendant's viewpoints, select one viewpoint that can form a point of dispute with the plaintiff's perspective.  &  SLC \\\midrule
2-9 & Event Detection: Given a sentence from a legal judgement document, detect which events are mentioned in this sentence.  &  MLC \\\midrule
2-10 & Trigger Word Extraction:Given a sentence from a legal judgment document and its corresponding events, predict which words in the sentence triggered these events. & Extraction  \\\midrule
3-1 & Fact-based Article Prediction:Given a fact statement from the legal judgement document, predict which article items should be applied. & MLC \\\midrule
3-2 & Scene-based Article Prediction: Given a described scenario and a related question, predict the corresponding article item. &  Generation \\\midrule
3-3 & Charge Prediction: Given fact statement from the legal judgement document and the applied article number, predict the cause of action (charge). & MLC \\\midrule
3-4 & Prison Term Prediction w.o. Article: Given fact statement from the legal judgement document, the applied article number and charge, predict the prison term. & Regression \\\midrule
3-5 & Prison Term Prediction w. Article:Given fact statement from the legal judgement document, the applied article content and charge, predict the prison term. &  Regression \\\midrule
3-6 & Case Analysis:Given a case and a corresponding question, select the correct answer from 4 candidates. &  SLC \\\midrule
3-7 & Criminal Damages Calculation:Given a fact description about a criminal process, predict the amount of money involved in this case. &  Regression \\\midrule
3-8 & Consultation:Given a user consultation, generate a suitable answer. &  Generation \\
\bottomrule
\end{tabularx}
}
\caption{Details of the definition and type of each task. LawBench contains 5 task types: generation, single-label classification (SLC), multi-label classification (MLC), regression, and extraction.}
\label{tab:table8}%
\end{table*}%
\endgroup

\textit{Knowledge Memorization} is the lowest cognitive level task in the benchmark. It examines the model's ability to accurately memorize laws and regulations as well as some legal knowledge. For detailed information, please refer to Table \ref{tab:table8} ID 1-1 and 1-2.

\textit{Knowledge Understanding} is the second level in the cognitive hierarchy. This category contains 10 tasks, detailed information can be found in Table \ref{tab:table8} ID 2-1 to 2-10.

\textit{Knowledge Application} is the most difficult category for the benchmark. There are 4 types of tasks in this level. First type is multi-label classification task. This kind of task includes 3-1(Fact-based article prediction) and 3-3(charge prediction).Then, 3-2(scene-based article prediction) and 3-8(consultation) are both generation task. Third one is regression task. 3-4(prison term prediction without article), 3-5(prison term prediction with article) and 3-7(criminal damages calculation) are all regression tasks. Last one 3-7(case analysis) is single-label classification task. Detailed information of these tasks can be found in Table \ref{tab:table8} ID 3-1 to 3-10.

\begin{table*}[htbp]
\centering
\small
 \resizebox{\columnwidth }{!}{
\begin{tabular}{l|ccccc|ccc|c}
\toprule
\multicolumn{1}{l|}{\multirow{3}*{Task}} &
\multicolumn{5}{c}{\shortstack{Legal-specific LLMs}} &
\multicolumn{3}{c}{\shortstack{General LLMs}} & \multicolumn{1}{c}{\shortstack{API-Based  LLMs}} \\
\multicolumn{1}{l|}{} & \multicolumn{1}{c}{\shortstack{InternLM-Law \cr 7B}} & \multicolumn{1}{c}{\shortstack{DISC-Law \cr7B}} & \multicolumn{1}{c}{\shortstack{Lawyer-LLaMA \cr13B}} & \multicolumn{1}{c}{\shortstack{ChatLaw \cr13B}} & \multicolumn{1}{c|}
{\shortstack{fuzimingcha \cr6B}} & \multicolumn{1}{c}{\shortstack{Qwen-1.5 \cr 7B}} & \multicolumn{1}{c}{\shortstack{InternLM2 \cr 7B}} & \multicolumn{1}{c|}{\shortstack{Qwen-1.5 \cr 72B}}
& \multicolumn{1}{c}{\shortstack{GPT-4 \cr N/A}} \\ 
\midrule
1-1 & \textbf{52.84} & 21.29 & 12.33 & 14.85 & 25.22 & 18.80   & 13.03 & 29.13 & 15.38\\
1-2 & 74.60 & 54.80 & 23.20 & 28.40 & 7.80 &  51.00  & 50.20 & \textbf{ 76.40} & 55.20 \\
\hline
2-1 & \textbf{57.27} & 12.23 & 4.33 & 12.22 & 4.93 & 12.00   & 36.78 &26.91& 12.53  \\
2-2  & \textbf{61.00} & 20.20 & 8.25 &2.68 & 19.59 & 31.80  & 39.20 & 48.60 & 41.65 \\
2-3  & \textbf{90.29} & 62.48 & 15.88 & 42.24 & 28.46 & 46.86 & 54.52 & 62.05 & 69.79  \\
2-4 & \textbf{49.00}  & 41.60 & 4.40 & 27.60 & 18.60 & 39.20  & 43.80 &39.00 & 44.00\\
2-5 & 87.38 & 60.20 & 34.61 & 39.11 & \textbf{97.59} & 62.57  & 47.21 & 66.47 & 56.50 \\
2-6 & 56.19 & 7.70 & 41.65 & 54.89 & 44.07 & 20.83  & 51.50 & \textbf{75.53} & 26.86 \\
2-7 & \textbf{53.21} & 33.71 & 38.51 & 38.45 & 54.32 & 30.59  & 33.60 & 34.81 & 37.92 \\
2-8 & \textbf{83.80}  & 27.20 & 9.60 & 18.60 & 8.80  & 38.40 & 43.20 & 54.40 & 61.20  \\
2-9  & \textbf{91.09}  & 84.89 & 29.78 & 31.74 & 16.90 & 58.65& 63.89 & 70.55 & 78.82  \\
2-10 & \textbf{88.89} & 14.08 & 2.38 & 14.56 & 7.78 & 16.37  & 36.32 & 43.29 & 65.09  \\
\hline
3-1 &  \textbf{75.59} & 43.96 & 0.60 & 33.28 & 25.19 & 57.53  & 63.79 &72.42 & 52.47 \\
3-2 & \textbf{47.82} & 38.70 & 25.94 & 31.55 & 22.18 & 31.93   & 14.12 & 29.67 & 27.54  \\
3-3 & \textbf{68.13 }& 50.21 & 31.30 & 27.90 & 55.93 & 45.35  & 48.91 & 57.07& 41.99  \\
3-4 & \textbf{84.22 } & 72.07 & 74.19 & 76.18 & 77.23 & 79.26 & 81.42 &81.32 & 82.62 \\
3-5  & 80.05 & 77.19 & 75.52 & 73.57 & 75.52 & 79.53 & 80.11 & 79.95 & \textbf{81.91}  \\
3-6 & 63.60 & 51.00 & 17.80 & 28.80 & 7.00 & 45.00  & 39.60 & \textbf{70.40} & 48.60 \\
3-7 & 66.00 & 42.80 & 39.20 & 41.40 & 47.20 & 43.00  & 55.40 & 74.80 & \textbf{77.60}  \\
3-8 & 23.17 & 15.63 & 16.94 & 17.17 & 16.64 & 19.51  & 19.32 & \textbf{24.30} & 19.65  \\
\hline
AVG & \textbf{67.71} &  41.60 & 25.32 & 32.76 &  33.05  &41.41  & 45.80 & 55.85 & 52.35  \\
\bottomrule
\end{tabular}
}
\caption{\small The zero-shot performance of the model on LawBench in each category.}
\label{table9}
\end{table*}

We provide zero-shot results on LawBench in Table \ref{table9} and one-shot results on LawBench in Table \ref{table10}.
\begin{table*}[!ht]
\centering
\small
 \resizebox{\columnwidth }{!}{
\begin{tabular}{l|ccccc|ccc|c}
\toprule
\multicolumn{1}{l|}{\multirow{3}*{Task}} &
\multicolumn{5}{c}{\shortstack{Legal-specific LLMs}} &
\multicolumn{3}{c}{\shortstack{General LLMs}} & \multicolumn{1}{c}{\shortstack{API-Based  LLMs}} \\
\multicolumn{1}{l|}{} & \multicolumn{1}{c}{\shortstack{InternLM-Law \cr 7B}} & \multicolumn{1}{c}{\shortstack{DISC-Law \cr7B}} & \multicolumn{1}{c}{\shortstack{Lawyer-LLaMA \cr13B}} & \multicolumn{1}{c}{\shortstack{ChatLaw \cr13B}} & \multicolumn{1}{c|}
{\shortstack{fuzimingcha \cr6B}} & \multicolumn{1}{c}{\shortstack{Qwen-1.5 \cr 7B}} & \multicolumn{1}{c}{\shortstack{InternLM2 \cr 7B}} & \multicolumn{1}{c|}{\shortstack{Qwen-1.5 \cr 72B}}
& \multicolumn{1}{c}{\shortstack{GPT-4 \cr N/A}} \\ 
\midrule
1-1  & \textbf{57.50} & 21.84 & 13.04 & 15.98 & 20.21 & 18.15   & 17.04 & 25.71 & 17.21\\
1-2  & \textbf{72.40} & 52.20 & 10.60 & 29.40 & 12.80 & 46.20   & 47.00 & 74.40 & 54.80 \\
\hline
2-1  &\textbf{ 57.27} & 13.44 & 4.90 & 13.01 & 2.86 & 14.51 & 36.78 & 35.01 & 18.31  \\
2-2 & \textbf{62.40}  & 21.40 & 19.20 & 9.00 & 2.40 &  22.80  & 40.00 & 44.20 & 46.00 \\
2-3  & \textbf{90.06} & 66.02 & 9.03 & 30.91 & 17.44 & 51.29  & 49.55 & 65.35 & 69.99  \\
2-4  & \textbf{49.00} & 42.80 & 3.00 & 26.60 & 8.80 & 40.00   & 41.80 & 40.60 & 44.40 \\
2-5 & 86.75 & 62.92 & 39.65 & 41.41 & \textbf{93.35 }& 64.6    & 61.61 & 78.46 & 64.80 \\
2-6 & 56.14 & 32.70 & 36.33 & 60.68 & 42.28 & 61.40  & 64.95& 73.83 & \textbf{79.96} \\
2-7 & \textbf{53.03}  & 25.16 & 37.10 & 42.41 & 31.43 & 33.47  & 37.12 & 42.11 & 40.52  \\
2-8 & \textbf{81.4} & 20.20 & 0.40 & 20.20 & 11.40 & 39.00   & 44.80 & 57.60 & 59.00  \\
2-9 & \textbf{91.39} & 81.60 & 33.19 & 40.27 & 21.26 & 62.96    & 66.54 & 74.71 & 76.55  \\
2-10  & \textbf{88.31} & 14.45 & 6.12 &17.37 & 7.04 & 22.41  & 40.18 & 37.35 & 65.26 \\
\hline
3-1 & \textbf{75.55} &65.61 & 0.33 & 25.99 & 3.86 & 53.57   & 64.15 & 73.79 & 52.20 \\
3-2 & \textbf{47.56} & 39.77 & 27.23 & 33.96 & 32.96 & 33.86   & 29.35 & 36.10 & 33.15  \\
3-3  & \textbf{68.62} &57.22 & 19.36 & 12.24 & 43.6 & 44.91 & 51.03 & 60.01 & 41.30  \\
3-4  & \textbf{83.83} & 75.41 & 70.99 & 74,31 & 78.95 & 80.86 & 80.11 & 80.77 & 83.21 \\
3-5  & 79.37 & 75.72  & 73.56 & 73.01 & 79 & 78.02  & 80.21 & 79.11 & \textbf{82.74}  \\
3-6  & 65.60 & 46.40 & 6.60 & 26.80 & 13.80 & 47.20 & 41.40 & \textbf{68.80} & 49.60 \\
3-7 & 64.80 & 51.20 & 33.80 & 42.00 & 38.20 & 42.40  & 56.60 & 75.00 & \textbf{77.00 } \\
3-8 & 22.37 & 13.76 & 16.02 & 16.72 & 13.95 &  19.84  & 20.42 & \textbf{24.67} & 19.90  \\
\hline
AVG   & \textbf{ 67.67} &  43.99 & 23.02 & 32.63 & 28.78 & 43.87   & 48.53 & 57.38 & 53.85  \\
\bottomrule
\end{tabular}
}
\caption{\small The one-shot performance of the model on LawBench in each category.}
\label{table10}
\end{table*}
\FloatBarrier

\definecolor{lightgray}{gray}{0.95}
\lstdefinestyle{prompt}{
    basicstyle=\ttfamily\fontsize{7pt}{8pt}\selectfont,
    frame=none,
    breaklines=true,
    backgroundcolor=\color{lightgray},
    breakatwhitespace=true,
    breakindent=0pt,
    escapeinside={(*@}{@*)},
    numbers=none,
    numbersep=5pt,
    xleftmargin=5pt,
}
\tcbset{
  aibox/.style={
    top=10pt,
    colback=white,
    colframe=black,
    colbacktitle=black,
    enhanced,
    center,
    % breakable,
    attach boxed title to top left={yshift=-0.1in,xshift=0.15in},
    boxed title style={boxrule=0pt,colframe=white,},
  }
}
\newtcolorbox{AIbox}[2][]{aibox, title=#2,#1}

\section{Applications} \label{AppendixC}

\paragraph{Legal Consultation.}
Consumer rights are a common topic of legal consultation, and our model can be a legal advisor to answer various questions in wide range of domains. For example, model can help you claim compensation for personal injury, give advice in employment contract, and protect your intellectual property. We illustrate our model's performance on a consumer rights question in Figure \ref{fig:legal_consultation}. In legal consultations, the user's queries are often vague and do not clearly articulate the facts of the case. Our model responds to this query by first listing possible scenarios, then attempting to analyze the legal issues in each scenario and identifying potentially relevant statutes. It then concludes by summarizing the possible scenarios.\\

\begin{figure*}[!ht] 
\vspace{-5mm}
\begin{AIbox}{Example of Legal Consultation}
{
    \textbf{Question:} \\
    %在餐厅吃饭，磕坏牙了，应该找谁赔偿？
        Can I claim compensation from the restaurant if I break teeth while eating there?\\
}
\\
\textbf{Model Answer:} \\

When dining at a restaurant and breaking teeth, you can consider the following points to determine who should be compensated:

\begin{enumerate}
    \item \textbf{Responsibility of the restaurant}: If you broke your teeth due to the presence of hard objects (such as bones, stones, etc.) in the food provided by the restaurant during normal dining, then the restaurant may be liable for compensation. 
    According to Article 148, Paragraph 2 of the "Food Safety Law of the People's Republic of China", in cases where food products that do not meet food safety standards are produced, or knowingly sold by businesses, consumers can demand compensation for losses and may also require the producer or operator to pay ten times the price of the products or three times the compensation for losses.
    \item \textbf{Providing evidence}: To protect your rights, you need to provide relevant evidence, such as retaining the food that caused your tooth damage, medical records, and expense receipts, to prove the causal relationship between your tooth damage and your dining experience at the restaurant.
    \item \textbf{Responsibility of third parties}: If your tooth damage is caused by the negligence of other customers or restaurant staff, then the responsibility may lie with that third party. In this case, you can first negotiate with the third party for a resolution, and if negotiation fails, you can consider legal recourse to protect your rights.
\end{enumerate}
If you believe the restaurant should be responsible for your tooth damage, you can first negotiate with the restaurant. If negotiation fails, you can seek legal assistance or file a complaint with the consumer association to protect your legal rights. Maintain calm and rationality during the problem-solving process, and retain relevant evidence.
\end{AIbox} 
\caption{An example of legal consultation.}
\label{fig:legal_consultation}
\end{figure*}
\FloatBarrier

\paragraph{Legal NLP Tasks.}
Our model is capable of dealing with various legal NLP task, including memorization, understanding and also apply legal knowledge. The tasks cover different types of NLP tasks, classification, extraction, regression and also generation.
We demonstrate our model's performance on a legal NLP task: predicting the charge of a crime based on the facts of the case. See Figure \ref{fig:nlp_task_example} for details. In this task, our model has a good understanding of the case and accurately analyzes the charges by utilizing corresponding legal knowledge. Furthermore, it will follow the instructions and provide answers that conform to the specified format.

\begin{figure*}[!h] 
\vspace{-5mm}
\begin{AIbox}{Example of Legal NLP Tasks }
{
    \textbf{Question:} \\
    Please simulate a judge and provide the charge based on the following fact. Only output the name of the charge, and place the answer between \textbf{[charge]} and \textbf{[eoa]}. For example, \textbf{[charge]} theft, scam \textbf{[eoa]}. Please follow the format strictly. \\
    Fact: The People's Procuratorate of Da Ming County accuses the defendant, Zhang Mou, of a crime. On March 25, 2014, at around 9 a.m., the defendant Zhang Mou got into a dispute with Ximou, a villager from his own village, at his home in Zhuangmou, due to certain reasons, which escalated into a physical fight. During the fight, the defendant Zhang Mou injured Ximou and his sons, Ximou B and Ximou C, using a crowbar. According to the appraisal by the Physical Evidence Appraisal Room of the Da Ming County Public Security Bureau, Ximou's injuries were classified as second-degree minor injuries, while Ximou B and Ximou C sustained minor injuries. To substantiate the aforementioned accusations, the prosecuting authority presented evidence including the victim's statement, the defendant's confession, witness testimonies, forensic appraisal reports on the extent of bodily injuries, on-site investigation records, and photographs. The prosecuting authority believes that the defendant Zhang Mou intentionally caused bodily harm to others, resulting in minor injuries, thus violating the provisions of the Criminal Law of the People's Republic of China. The facts of the crime are clear, and the evidence is reliable and sufficient, warranting criminal liability under the relevant provisions. Additionally, the plaintiff and their legal representative in the accompanying civil litigation do not contest the characterization of this case and request the defendant to compensate for the economic losses incurred, totaling 120,960.6 yuan, due to their actions. They have submitted relevant evidence, including medical bills, appraisal fees, diagnosis certificates, and hospitalization records.\\

}
\textbf{Model Answer:} \\
\textbf{[charge]}intentional injury\textbf{[eoa]}
\end{AIbox} 
\caption{An example of legal NLP Tasks.}
\label{fig:nlp_task_example}
\end{figure*}
\FloatBarrier

\paragraph{Tool Usage.}
Our training strategy not only enables the model to retain its existing capabilities but also facilitates the integration of general competencies with newly acquired legal expertise, thereby enhancing the model's performance on legal tasks. We tested the model by writing code to solve amount calculation problems in LawBench. Detailed examples are shown in Figure \ref{fig:tooL_use}. It is worth noting that none of the data points in our training dataset are examples of solving amount calculation problems using code. And by using programming methods, we find that model can improve its accuracy in the task compared generate its final output directly. As our example demonstrates, our model successfully wrote a code snippet to solve the criminal damages calculation problem. Model accurately extracted the relevant amounts from the text and wrote executable Python code, demonstrating the model's ability to perform numeric computations.

\FloatBarrier

\FloatBarrier
\begin{figure*}[!h] 
\vspace{-5mm}
\begin{AIbox}{Example of Tool Usage}
{
    \textbf{Question:} \\
    Please write a code to calculate the amount of money involved in the following case: The public prosecution accuses the defendant Meng of stealing RMB 1400 and an Apple iPhone 4 mobile phone valued at RMB 1300 from the victim Xiao's pocket and computer desk, respectively, while Xiao was washing up in the bathroom of Room 205 at the "Xiangyun Hotel" at 114 Gongnong Road, Dadong District, Shenyang City at around 8 am on July 14, 2013. The stolen money has been spent. Meng was arrested by the public security organs on July 19, 2013. The above facts were not objected to by defendant Meng during the court trial. There is evidence such as proof of arrest, physical evidence photos, detention decision, detention list, return list, basic information of permanent residents, phone search records, and identification records, which are sufficient to establish the case.\\
}
\\
\textbf{Model Answer:} \\
According to the information provided, the following Python code can be written to calculate the involved amount:
\begin{lstlisting}
# Involved amount
amount_1 = 1400
amount_2 = 1300

# Total amount involved
total_amount = amount_1 + amount_2

# Print total amount involved
pirnt("Total amount involved: ", total_amount)
\end{lstlisting}

\end{AIbox} 
\caption{An example of Tool Usage.}
\label{fig:tooL_use}
\end{figure*}
\FloatBarrier

\end{document}